%% file: main.tex
\pdfoutput=1

\documentclass[11pt]{article}

\usepackage[final]{acl}

\usepackage{times}
\usepackage{latexsym}

\usepackage[T1]{fontenc}

\usepackage[utf8]{inputenc}

\usepackage{microtype}

\usepackage{inconsolata}

\usepackage{amsmath}
\usepackage{amsfonts}
\usepackage{graphicx}
\usepackage{booktabs}
\usepackage{subcaption}
\usepackage{multirow}
\usepackage{adjustbox}

\newcommand{\xprompt}{x^{\mathrm{prompt}}}
\newcommand{\xdum}{x^{\mathrm{dum}}}
\newcommand{\xdock}{x^{\mathrm{doc}}_{k}}
\newcommand{\xdocki}{x^{\mathrm{doc}}_{k,i}}
\newcommand{\xdoc}{x^{\mathrm{doc}}}
\newcommand{\xdocone}{x^{\mathrm{doc1}}}
\newcommand{\xdoctwo}{x^{\mathrm{doc2}}}

\newcommand{\rel}{\mathrm{rel}}
\newcommand{\bias}{\mathrm{bias}}

\newcommand{\Attn}{\mathrm{Attn}}

\newcommand{\Attncal}{\mathrm{Attn_{calibrated}}}
\newcommand{\Attnori}{\mathrm{Attn_{original}}}
\newcommand{\attnori}{\mathrm{attn_{original}}}
\newcommand{\attncal}{\mathrm{attn_{calibrated}}}

\newcommand{\eq}{Eq.}

\newcommand{\vanilla}{\text{Vanilla attention}}
\newcommand{\querygen}{\text{Query generation}}
\newcommand{\relgen}{\text{Relevance generation}}
\newcommand{\llmlinguafull}{\text{LongLLMLingua-$r_k$}}
\newcommand{\llmlingua}{\text{LongLLMLingua-$r_k$}}

\newcommand{\vicunafull}{\texttt{Vicuna-7b-v1.5-16k}}

\title{Found in the Middle: Calibrating Positional Attention Bias\\Improves Long Context Utilization}

\author{
Cheng-Yu Hsieh$^{1}$\thanks{Work done while the author was a student researcher at Google Cloud AI Research. Correspondence to: Cheng-Yu Hsieh <cydhsieh@cs.washington.edu>, Chen-Yu Lee <chenyulee@google.com>}, \
Yung-Sung Chuang$^{2}$, \
Chun-Liang Li$^{3}$, \
Zifeng Wang$^{3}$, \\
\bf
Long T. Le$^{3}$, \
Abhishek Kumar\thanks{Work done while the author was at Google DeepMind.}, \
James Glass$^{2}$, \
Alexander Ratner$^{1}$, \\
\bf
Chen-Yu Lee$^{3}$, \
Ranjay Krishna$^{1}$\thanks{The authors contributed equally to this work.}, \
Tomas Pfister$^{3}$\footnotemark[3]\\
$^1$University of Washington,
$^2$MIT,
$^3$Google Cloud AI Research
}

\begin{document}
\maketitle
\begin{abstract}

\input{sections/00-abstract}
\end{abstract}

\input{sections/01-intro}
\input{sections/02-attention_bias}

\input{sections/03-calibrated_attention}

\input{sections/04-experiment}

\input{sections/05-related_work}

\input{sections/06-conclusion}

\newpage
\bibliography{main}

\newpage
\appendix

\input{sections/07-appendix}

\end{document}

%% file: sections/00-abstract.tex
Large language models (LLMs), even when specifically trained to process long input contexts, struggle to capture relevant information located in the middle of their input. 
This phenomenon has been known as the \textit{lost-in-the-middle} problem.
In this work, we make three contributions.
First, we set out to understand the factors that cause this phenomenon.
In doing so, we establish a connection between lost-in-the-middle to LLMs' intrinsic attention bias: LLMs exhibit an \textit{U-shaped attention bias} where the tokens at the beginning and at the end of its input receive higher attention, regardless of their relevance.
Second, we mitigate this positional bias through a calibration mechanism, \textit{found-in-the-middle}, that allows the model to attend to contexts faithfully according to their relevance, even though when they are in the middle.
Third, we show \textit{found-in-the-middle} not only achieves better performance in locating relevant information within a long context, but also eventually leads to improved retrieval-augmented generation (RAG) performance across various tasks, outperforming existing methods by up to 15 percentage points. These findings open up future directions in understanding LLM attention bias and its potential consequences.

%% file: sections/01-intro.tex
\begin{figure}[t]
    \centering
    \includegraphics[width=\linewidth]{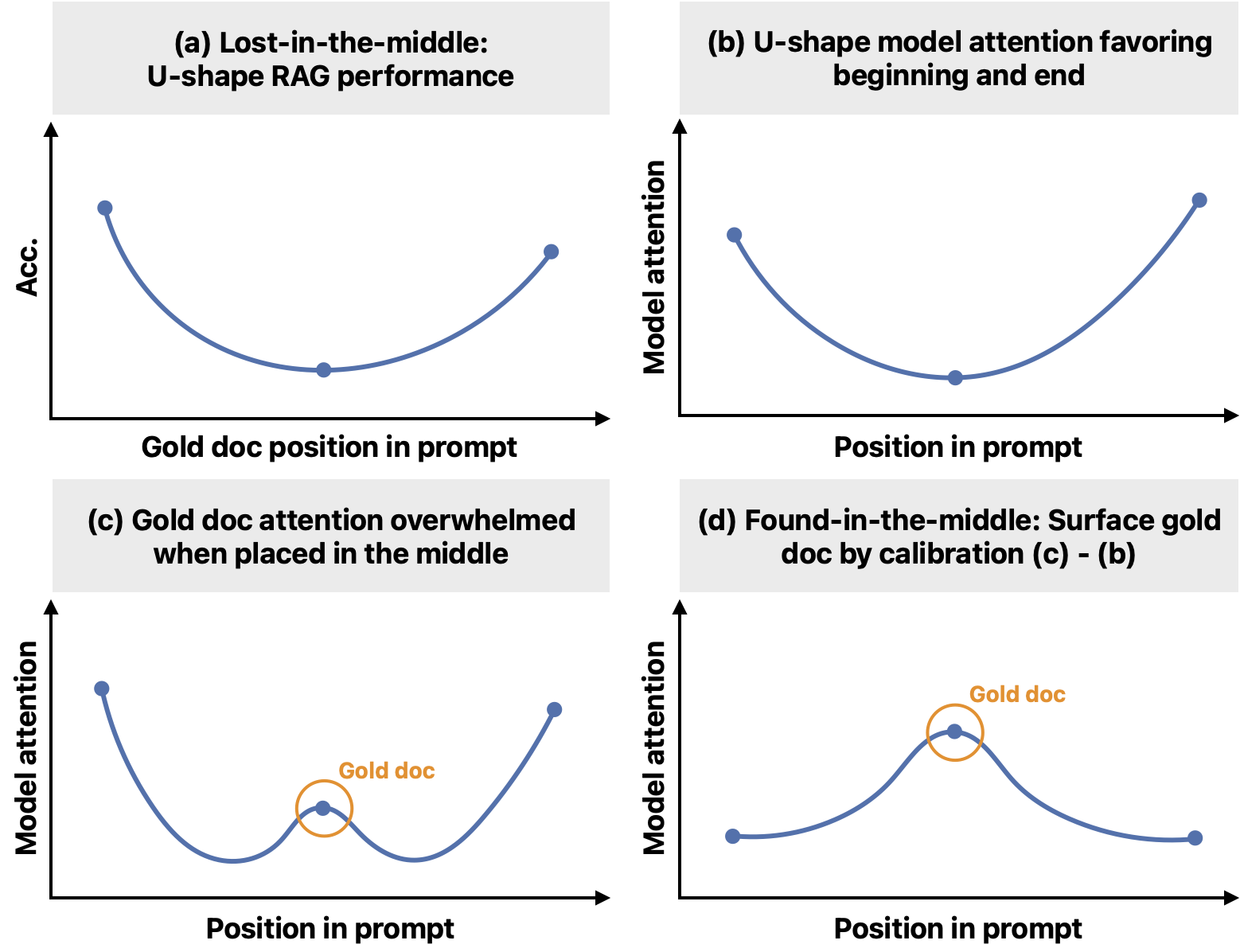}
    \caption{(a) Lost-in-the-middle refers to models' U-shape RAG performance as the relevant context's (e.g., a gold document containing the answer to a query) position varies within the input; (b) We observe models exhibit U-shape attention weights favoring leading and ending contexts, regardless of their actual contents; (c) Models do attend to relevant contexts even when placed in the middle, but are eventually distracted by leading/ending contexts; (d) We propose a calibration mechanism, found-in-the-middle, that disentangles the effect of U-shape attention bias and allows models to attend to relevant context regardless their positions.}
    \label{fig:intro}
\end{figure}

\section{Introduction}
Effective prompting of large language models (LLMs)~\citep{brown2020language,anil2023palm,touvron2023llama} has enabled a variety of user-facing
applications, including conversational interfaces (chatbots)~\cite{thoppilan2022lamda}, search and summarization~\cite{silo}, open-domain question answering~\citep{izacard2021leveraging}, tool usage~\citep{hsieh2023tool}, fact checking~\citep{asai2023self}, and collaborative writing~\cite{lee2019latent}.
Some of these applications, such as search and summarization~\citep{ji2023survey,factscore,asai2023self}, require the ability to retrieve information from external knowledge sources.
As a result, retrieval-augmented generation (RAG) has become a powerful solution. RAG fetches relevant documents (e.g.~structured tables~\citep{wang2024chain} and API documentation~\citep{karpukhin2020dense}) from external knowledge sources and makes them available in the LLMs' input prompt~\citep{Khandelwal2020Generalization, borgeaud2022improving, izacard2022few, xu2023retrieval}.
Despite the widespread utility of RAG~\citep{longchat2023,xiong2023effective,chatgpt,team2023gemini}, recent experiments highlight a striking deficiency: LLMs struggle to locate relevant documents when they are placed in the middle of their input prompts~\citep{liu2023lost,longchat2023}.
They call this the \textit{lost-in-the-middle} phenomenon.

To overcome this phenomenon, a few mechanistic strategies have been proposed~\citep{jiang2023longllmlingua, peysakhovich2023attention}.
These methods \textit{re-rank} the relevance of different documents and \textit{re-order} the most relevant ones to either the beginning or end of the input context.
Unfortunately, re-ranking usually requires additional supervision or dedicated finetuning for performant RAG performance~\citep{karpukhin2020dense, shi2023replug, sun2023chatgpt}. Worse, re-ranking methods do not fundamentally improve LLMs' ability to utilize and capture relevant information from the provided input contexts.
The underlying causes of this behavior remains unclear, even though it has been observed across multiple decoder-only LLMs~\citep{touvron2023llama, longchat2023, chatgpt}.

In this work, we make three contributions: 
First, we set out to understand the potential factors leading to the \textit{lost-in-the-middle} problem.
\textbf{We establish a connection between lost-in-the-middle to LLMs' intrinsic attention bias} (see Figure~\ref{fig:intro}).
Specifically, we find that models often demonstrate a \textit{U-shaped} attention distributions, with higher attention values assigned to the beginning and end of the input prompt. This correlates well with the U-shaped RAG performance observed in prior literature~\citep{liu2023lost}.
Interestingly, this focus on the beginning and end also extends to content utilization: models preferentially use information from the beginning and end of their prompts~\citep{ravaut2023position,peysakhovich2023attention}.  
This leads us to hypothesize that the positional attention bias may contribute to the phenomenon, wherein the bias could lead to over-reliance on content at the beginning/end of the input, regardless of its true relevance.

Second, we verify our hypothesis by intervening on this attention bias to determine its impact on performance.
\textbf{We propose a mechanism to disentangle positional bias from model's attention.}
We first esitmate this bias through measuring the change in attention as we vary the relative position of a fixed context in the LLM's prompt.
By quantifying and then removing this bias from the attention scores for a given query, we can obtain the \textit{calibrated attention} scores across the retrieved documents. 
This calibrated attention proves to be better correlated to the ground truth relevance of the document to a user query. 
In open-domain question answering tasks~\citep{kwiatkowski2019natural}, our proposed calibrated attention outperforms popular existing approaches for ranking the relevance of retrieved documents (up to 48 Recall@3 points).
This finding challenges the recent belief that LLMs struggle to capture relevant context embedded in the middle of inputs, suggesting they may indeed be capable of doing so, but are only hindered by the overwhelming positional bias.

Third, we operationalize our calibration mechanism as a solution for this phenomenon, naming our attention intervention \textit{found-in-the-middle}.
\textbf{We show that calibrating the attention leads to improvements across two popular LLMs with different context window lengths on two RAG tasks.}
Our experiments demonstrate improvements over standard model generation by up to 15 percentage point on NaturalQuestion dataset~\cite{kwiatkowski2019natural}.
We hope the work opens up future directions in understanding LLM's attention biases and their effect on downstream tasks.

%% file: sections/02-attention_bias.tex
\begin{figure*}[t!]
    \centering
    \includegraphics[width=\linewidth]{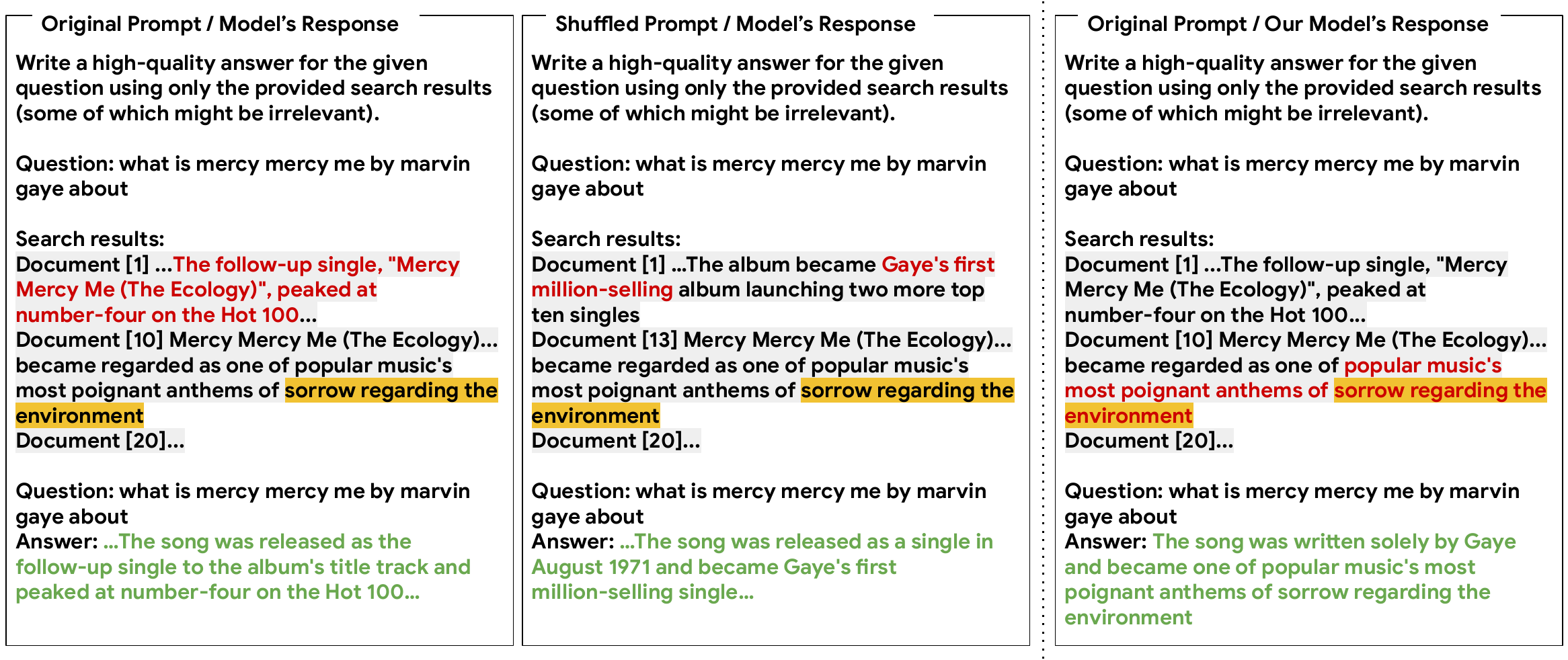}
    \caption{\textbf{Left and Middle: Qualitatively, the model's response exhibits a strong bias towards the document at the first position (red).} This persists whether the input documents retain their original order (left: gold document at the 10th position) or are randomly shuffled (middle: gold document at the 13th position). Model responses are shown in green, with the gold answer highlighted in yellow. \textbf{Right: Our attention calibration method enables the model to find relevant context even when placed in the middle.}}
\label{fig:generation_bias_qual}
\end{figure*}

\begin{figure}[t]
    \centering
    \includegraphics[width=0.98\linewidth]{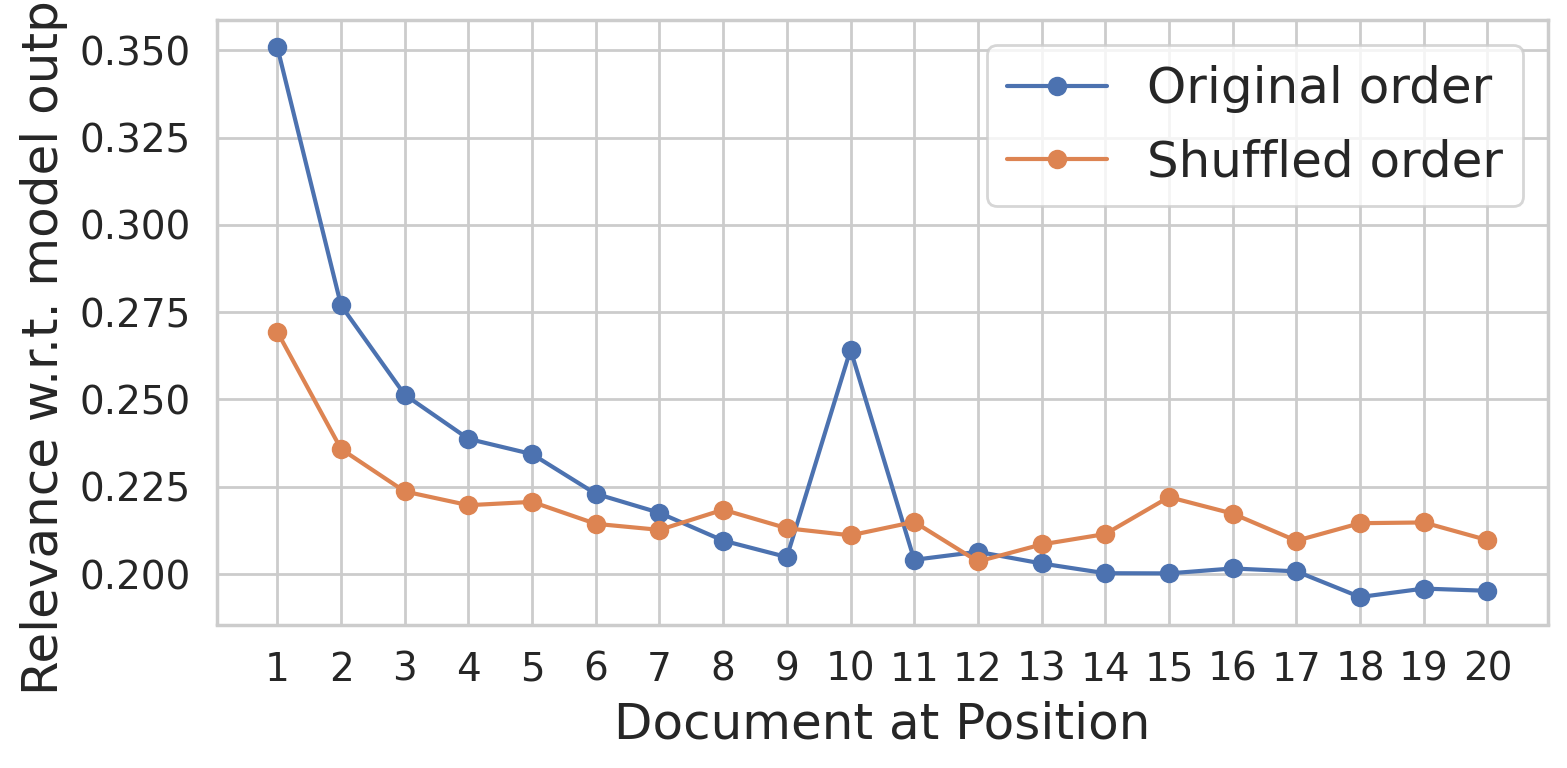}
    \caption{\textbf{Quantitatively, the model's response strongly depends on the document at the first position.} This dependence persists even after randomly shuffling the document order, irrespective of its relevance to the query. We measure this dependence by computing the TF-IDF similarity score between the response and each document (gold document originally at position 10).}
    \label{fig:generation_bias_quant}
\end{figure}

\begin{figure}[t]
    \centering
    \includegraphics[width=0.98\linewidth]{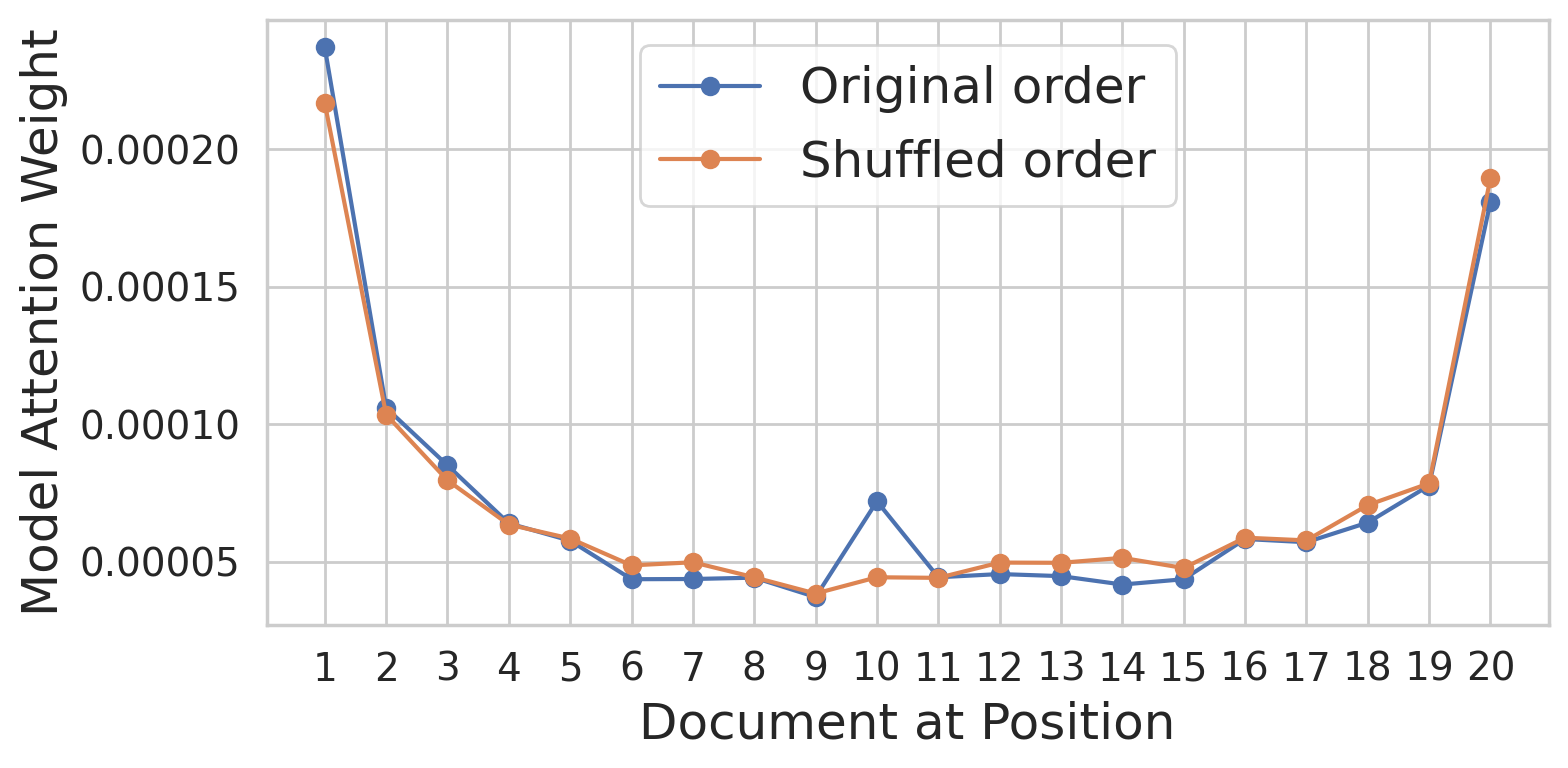}
    \caption{\textbf{Average attention weights reveal a U-shaped positional bias in the model.}  Documents at the beginning and end receive greater attention,  regardless of order (gold document originally at position 10). Attention is averaged across different decoder layers and attention heads.}
    \label{fig:attn_bias}
\end{figure}

\section{Positional attention bias overpowers mid-sequence context}
\label{sec:preliminary}

Recent work has produced language models capable of handling increasingly long input contexts~\citep{ xiong2023effective, longchat2023}. However, many of these models struggle to locate relevant information placed in the middle of the input sequence~\citep{liu2023lost}, a phenomenon known as the ``lost-in-the-middle'' problem.
While this problem is widely recognized, the potential factors contributing to this behavior remain poorly understood. In this work, we seek to deepen our understanding of the problem through a suite of exploratory qualitative and quantitative studies.

\paragraph{Setup.} We adhere to the original experimental setup outlined in \citet{liu2023lost}, utilizing an open-domain question answering task~\citep{kwiatkowski2019natural} for our exploratory study. In the lost-in-the-middle setup~\citep{liu2023lost}, a model is tasked to answer a user query $x^\mathrm{q}$ using a set of $k$ related documents retrieved from an external data source $D = \{x^{\mathrm{gold}}, x^{\mathrm{distract}}_1, \ldots, x^{\mathrm{distract}}_{k-1}\}$, where only the gold document $x^{\mathrm{gold}}$ contains the correct answer. The question and documents are typically serialized as an input sequence $x^{\mathrm{prompt}} = [x^\mathrm{q}, x^{\mathrm{doc}}_1, ..., x^{\mathrm{doc}}_k, x^\mathrm{q}]$, prompting a language model to generate the final answer\footnote{We repeat the question before and after the documents so that the model can better attend to relevant contexts~\citep{liu2023lost, xu2023re}.}. Observations indicate that model performance significantly decreases when $x^{\mathrm{gold}}$ is placed within the middle of the input prompt (i.e., $x^{\mathrm{doc}}_{{\lfloor}{k/2}{\rfloor}}$), compared to scenarios where $x^{\mathrm{gold}}$ is placed at the beginning or end.
Here, we reproduce lost-in-the-middle phenomenon with a \vicunafull\ (Vicuna) model~\citep{longchat2023} to gain deeper insights into the characteristics of the model's errors.
We focus our error analysis on the setting where we have a total of 20 documents ($K =20$). We specifically look at the examples where the model makes incorrect predictions when the gold document is placed at the middle (10-th) position.

\subsection{U-shaped attention bias}
We first examine responses generated when gold documents are placed in the \textbf{middle} of input prompts. Qualitatively, the model's response exhibits a strong bias towards the document at the first position, regardless of the gold document's location (Figure~\ref{fig:generation_bias_qual}). This bias persists whether the input documents retain their original order or are randomly shuffled.

The strong correlation between the model's output and the first document could suggest that they are highly relevant, distracting the model~\citep{shi2023large}.
However, quantitatively, the model's response strongly depends on the document at the first position (Figure~\ref{fig:generation_bias_quant}). This dependence persists even after randomly shuffling the document order, irrespective of its relevance to the query. We measure the dependence by computing the TF-IDF similarity between the response and each document (gold document originally at position 10).

To investigate the potential origins of positional bias, we visualize the model's self-attention weights, as the weights has been shown to correlate with models' generations, although not necessarily causal ~\citep{dong2021fly, zhang2023tell}.
More formally, given an input prompt consisting of $K$ documents $x^{\mathrm{prompt}} = [x^{\mathrm{doc}}_1, ..., x^{\mathrm{doc}}_K]$, where each document $x^{\mathrm{doc}}_k = \{x^{\mathrm{doc}}_{k, i}\}^{N_k}_{i=1}$ contains $N_{k}$ tokens,
let $\mathrm{Attn}: \mathcal{X} \times \mathbb{N} \to \mathbb{R}$ denote a function that computes the average attention weights assigned to document $x^{\mathrm{doc}}_k$ as $\mathrm{Attn}(x^{\mathrm{prompt}}, k) = \sum_{i=1}^{N_k} \mathrm{attn}(x^{\mathrm{doc}}_{k, i})/N_k$, where $\mathrm{attn}(x^{\mathrm{doc}}_{k, i})$ is the attention weight value allocated to token $x^{\mathrm{doc}}_{k, i}$ when predicting the next $|x^{\mathrm{prompt}}| + 1$ token.

Specifically, we visualize the self-attention weights assigned to each document, averaged across all its tokens, all decoder layers, and heads. We investigate how these weights vary based on document position within the input prompt. Interestingly, Figure~\ref{fig:attn_bias} (blue curve) reveals a U-shaped attention pattern. Documents near the beginning and end of the input receive higher weights, while those in the middle receive lower weights. Crucially, the U-shaped pattern persists even after randomly shuffling document order (Figure~\ref{fig:attn_bias}, orange curve), suggesting that this bias does not depend on the documents' actual content.

\subsection{Does attention favor relevant context?}

\paragraph{Observation 1: Model prioritizes relevant contexts from the same position.}
In Figure~\ref{fig:attn_bias}, we observe a significant difference in attention values at $x^{\mathrm{doc}}_{10}$ when comparing examples with original document order (blue) and randomly shuffled order (orange). Specifically, the attention value is notably higher when when $x^{\mathrm{doc}}_{10}$ is controlled to be $x^{\mathrm{gold}}$. This contrasts with instances where $x^{\mathrm{doc}}_{10}$ is uncontrolled, suggesting that apart from U-shaped positional bias, the model exhibits an ability to \textit{prioritize} relevant context.

\paragraph{Observation 2: Model prioritizes highly-weighted documents for generation.}
Based on these observations, we hypothesize that  positional attention bias significantly influence  the model's tendency to rely heavily on the first documents during output generation. Specifically, the models are more likely to incorporate the document receiving the highest attention (often the first) into its output. 
To validate this, for each of the examples of interest (where the model makes incorrect predictions), we divide their documents into first half receiving higher model attention and second half receiving lower attention. We then count the number of examples in which the first or second half contains the document that is most likely used in the model's generation (i.e., having the highest TF-IDF score with model's response).
In Table~\ref{table:contigency_table}, we show that documents receiving higher attention positively correlates with them being used in the model's generation.

From the above studies, we see that not only the model exhibits a U-shape positional attention bias, but this bias also correlates strongly with the model's biased tendency in using documents placed at certain positions in forming its response. We thus conjecture that lost-in-the-middle happens because of the dominating force of positional bias.

\begin{table}[t]
\caption{Number of examples where the most likely used document in the model's generation falls within the first half of documents receiving higher model attention or second half receiving lower attention. We see that there is a strong correlation where documents receiving higher attention are more likely to be used in model's response.}
\label{table:contigency_table}
\small
\centering
\begin{tabular}{lccc}
\toprule
 & \multicolumn{2}{c}{Most Likely Used} \\
 \cmidrule{2-3}
 & \# of examples & \%\\
\midrule
Highest Half Attention & $526$ & $74\%$ \\
Lowest Half Attention & $186$  & $26\%$ \\
\bottomrule
\end{tabular}
\end{table}

%% file: sections/03-calibrated_attention.tex
\section{Found-in-the-middle: modeling and isolating positional attention bias}
\label{sec:calibrated_attention}

Ideally, a model should leverage contexts in the input prompts---faithfully according to their relevance---for generating the response, instead of biasing towards contexts placed at certain positions within the input.
Towards this goal, we are interested in modeling the positional attention bias and mitigating it such that model attention can reflect the true relevance of the input context and ultimately improve models' effective utilization of the full context window.

\subsection{Two main factors in model attention}
In Sec.~\ref{sec:preliminary}, we find that there are two main forces driving the model attention assigned to different documents of an input prompt: (a) where the document locates within the entire input, and (b) the relevance of the document.

\paragraph{Our hypothesis.} We thus consider modeling the observable attention weights allocated to the $k$-th document of an input $x^{\mathrm{prompt}}$ as:
\begin{equation}
\label{eq:main-hypothesis}
    \mathrm{Attn}(x^{\mathrm{prompt}}, k) = f(\mathrm{rel}(x^{\mathrm{doc}}_k), \mathrm{bias}(k)), %
\end{equation}
where $\mathrm{rel(\cdot)}$ measures the relevance of an input document, $\mathrm{bias}(\cdot)$ characterizes the positional attention bias, and $f(\cdot)$ is some unknown monotonically increasing function w.r.t. to both $\mathrm{rel}(x^{\mathrm{doc}}_k)$ and $\mathrm{bias}(k)$. For ease of exposition, in the remainder of the paper, we overload $\Attn({\xdoc, k})$ to denote the attention value assigned to document $\xdock$ placed at the $k$-th position within an input prompt containing $K$ documents.

\paragraph{Corroborating our assumed model.}
Here, we conduct a suite of controlled experiments using NaturalQuestion with $K=20$ and a \vicunafull\ model to corroborate our assumed model. Specifically, for Eq.~\ref{eq:main-hypothesis} to hold, it implies that:

\noindent\textbf{Condition~1:} When the relevance term is fixed, model attention increases as positional bias increases. That is, given two documents $\xdocone$ and $\xdoctwo$: \textit{if $\Attn(\xdocone, k) > \Attn(\xdocone, l)$, then $\Attn(\xdoctwo, k) > \Attn(\xdoctwo, l)$}.

\noindent\textbf{Condition~2:} Similarly, when the document position $k$ is fixed, model attention increases as the relevance of the document increase: \textit{if $\Attn(\xdocone, k) > \Attn(\xdoctwo, k)$, then $\Attn(\xdocone, l) > \Attn(\xdoctwo, l)$}.

We validate Condition~1 and~2 on $100$ randomly sampled examples from NaturalQuestion dataset, each with $K=20$ documents. For validating Condition~1, given a pair of documents $(\xdocone, \xdoctwo)$ and positions $(k, l)$, we can compute whether the relationship holds across all possible pairs. We can similarly test for Condition~2.
In Table~\ref{table:rank_corr}, we see that the percentage of valid example pairs are decently high, $83\%$ and $72\%$ respectively, for both conditions, providing supports to our hypothesis.

\begin{table}[t]
\caption{High correlations between model attention with document relevance and positional bias supports our hypothesized model.}
\label{table:rank_corr}
\small
\centering
\begin{adjustbox}{width=1\linewidth}
\begin{tabular}{lllc}
\toprule
Hypothesis test & $\rel(\xdoc)$ & $\bias(k)$ &  \% of valid pairs  \\
\midrule
Condition 1 & Fixed & Varying & 83\% \\
Condition 2 & Varying & Fixed & 72\% \\
\bottomrule

\end{tabular}
\end{adjustbox}
\end{table}

Recall that our goal is to disentangle positional attention bias from model attention such that the model can faithfully attend to relevant contexts, independent from their positions. So far, while we have established the monotonic increasing nature of $f$ in \eq~\ref{eq:main-hypothesis}, we have yet characterize the actual form of $f$ to remove the positional bias term from model attention.

To approximate $f$, we consider simple linear models by following machine learning principles (a.k.a. Occam's razor), for robust estimation: 
\begin{equation}
\label{eq:add}
    \Attn(\xdoc, k) = \rel(\xdoc) + \bias(k) + \epsilon,
\end{equation}
where $\epsilon$ is a noise.

To test how the model captures the underlying relationship, we compute Spearman's rank correlation between $\Attn(\xdocone, k) - \Attn(\xdoctwo, k)$ and $\Attn(\xdocone, l)  - \Attn(\xdoctwo, l))$ over quadruplets of $(\xdocone, \xdoctwo, k, l)$ collected from NaturalQuestion. A high correlation indicates 
small discrepancy between $\Attn(\xdocone, k) - \Attn(\xdoctwo, k)$ and $\Attn(\xdocone, l)  - \Attn(\xdoctwo, l))$. From our study, the linear model results in decently high correlation, $0.76$, suggesting its effectiveness despite the simplicity. 
We therefore adopt \eq~\ref{eq:add} as our model and leave other alternatives with more degree of freedoms as future work~\footnote{In Appendix~\ref{sec:addition_exp}, we also explore log-linear models, which  results in competitive $0.75$ rank correlation.}.

\subsection{Disentangling positional attention bias}
Most notably, having a simple form of $f$ allows us to isolate the effect of positional bias from model attention. Specifically, following from \eq~\ref{eq:add}, we can first obtain a reference model attention value with a dummy document $\xdum$ by:
\begin{equation}
\label{eq:dummy}
    \Attn(\xdum, k) = \rel(\xdum) + \bias(k) + \epsilon.
\end{equation}
By subtracting \eq~\ref{eq:add} and \eq~\ref{eq:dummy}, we can offset the bias term and obtain:
\begin{align}
\label{eq:offset}
    & \rel(\xdoc) \\
    & = \Attn(\xdoc, k) - \Attn(\xdum, k) + \rel(\xdum) \nonumber
\end{align}
Consider using a consistent dummy document $\xdum$ which has a constant $\rel(\xdum)$, we are then able to obtain the true relevance of different documents $\xdoc$, free from the positional bias. We refer to $\Attn(\xdoc, k) - \Attn(\xdum, k)$ as \textit{calibrated attention} as it removes the baseline attention, and call the overall calibration mechanism \textit{found-in-the-middle}.

\paragraph{Calibrated attention finds relevant contexts in the middle.} \eq~\ref{eq:offset} allows us to leverage calibrated attention to estimate and rank the relevance of different documents within an input prompt. To validate the effectiveness of our model, we evaluate using calibrated attention to re-rank documents in an input prompt w.r.t. a given query. We evaluate on NaturalQuestion with the Vicuna model where we focus on the most challenging setting when the gold document in placed in the middle of the input prompt. We compare our model to:
\begin{itemize}
    \item \vanilla: Using uncalibrated attention $\Attn(\xprompt, k)$ to rank the documents.
    \item \querygen~\citep{sun2023chatgpt}: Using likelihood of the model in generating the query based on the document.
    \item \relgen~\citep{sun2023chatgpt}: Prompting the model to answer whether a document is relevant to a query.
\end{itemize}

In Table~\ref{table:ranking_exp}, we compare Recall@3 of different methods where we vary the total number of documents retrieved. We see that the proposed calibrated attention consistently outperforms vanilla attention by a large margin, and also shows superior performances when compared to the other two re-ranking metrics.
The results validate that our proposed modeling approach is effective, and that if calibrated appropriately, language models can locate relevant information even when they are hidden in the middle of the input. 

\begin{table}[!t]
\caption{Calibrated attention outperforms existing methods in ranking the relevance of retrieved contexts given a user query. We report Recall@3 on NaturalQuestion when gold documents are placed in the middle of input context.}
\label{table:ranking_exp}
\small
\centering
\begin{tabular}{l cc}
\toprule
& \multicolumn{2}{c}{Number of total documents} \\
\cmidrule{2-3} 
Method & $K=10$ & $K=20$ \\
\midrule

\vanilla & 0.3638 & 0.2052 \\
\querygen & 0.6851 & 0.5815 \\
\relgen & 0.5521 & 0.4012  \\
Calibrated attention & {\bf 0.7427} & {\bf 0.6832}\\

\bottomrule
\end{tabular}
\end{table}

%% file: sections/04-experiment.tex
\section{Improving long-context utilization with found-in-the-middle}
Having validated that calibrated attention through found-in-the-middle is effective in locating relevant information within a long input context, we are ultimately interested in leveraging it to tackle lost-in-the-middle problem and practically improve a model's RAG performance.

\subsection{Attention calibration}
To allow the model to attend to contexts without being dictated by positional bias, we propose to intervene the model's attention based on the proposed calibrated attention. Specifically, given an input $\xprompt$, instead of allocating $\rel(\xdock) + \bias(k)$ attention to the $k$-th document, our ideal model attention $\Attncal(\xdock)$ would reflect only the relevance of the context $\rel(\xdock)$.

To achieve this, we propose to redistribute the attention values assigned to $\{\xdock\}_{k=1}^K$ according to $\rel(\xdock)$. Specifically, for each document $\xdock$, we propose to rescale the attention values on the tokens within the document, $\{\xdocki\}_{i=1}^{N_k}$, by:
\begin{align}
\label{eq:intervene}
    & \attncal(\xdocki) = \\
    & \frac{\alpha_k }{\Attnori(\xdock)}  \cdot \attnori(\xdocki) \cdot C, \nonumber
\end{align}
where $\alpha_k = \mathrm{Softmax}(\rel(\xdock), t)$, $t$ is the temperature hyperparamter, and $C$ is a normalization constant to ensure the total attention $\sum_{k,i}\xdocki$ remains unchanged. 
With the rescaling, we effectively make the final attention on $\xdock$:
\begin{equation}
    \Attncal(\xdock) \propto \mathrm{Softmax}(\rel(\xdock), t),
\end{equation}
where higher attention is allocated to more relevant context, and $t$ controls the disparity level.

\begin{figure*}[t!]
    \centering
    \includegraphics[width=\linewidth]{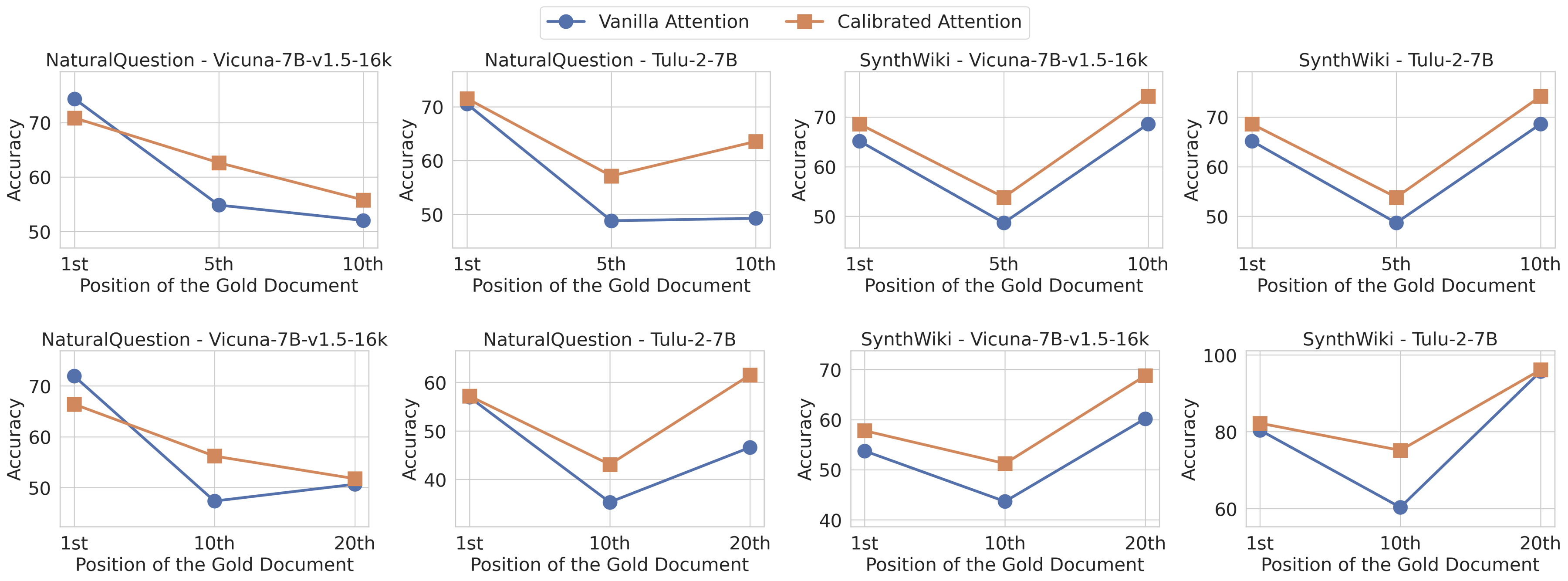}
    \caption{\textbf{Attention calibration effectively improves models' context utilization ability, with its performance curves lying almost entirely above standard vanilla attention (on 22 out of 24 cases). On the most challenging settings where the gold documents are placed in the middle, attention calibration provides 6-15 points improvements.} Top/Bottom row: 10/20-doc. Numbers shown in Table~\ref{table:rag_table}.}
\label{fig:baseline}
\end{figure*}

\subsection{Calibrated v.s. uncalibrated attention}
We evaluate the performance of the proposed attention calibration method. We conduct experiments on two multi-document question answering tasks (more details in Appendix~\ref{sec:datasets}), NaturalQuestion~\citep{kwiatkowski2019natural} and SynthWiki~\citep{peysakhovich2023attention}, with two models supporting different context window length:
\vicunafull\ (Vicuna)~\citep{longchat2023} and \texttt{tulu-2-7b} (Tulu)~\citep{wang2023far} with 16k and 8k context window respectively. For each dataset, we consider two settings with different number of retrieved documents, $K=\{10, 20\}$. We leave further implementation details in Appendix~\ref{sec:exp_detail}.

\paragraph{Found-in-the-middle improves long-context utilization across various datasets and models.}
In Figure~\ref{fig:baseline}, we see that found-in-the-middle attention calibration consistently outperforms the uncalibrated baseline by a large margin (up to 15 percentage point (pp) improvement) across different tasks and models. On the most challenging scenario when the gold document is placed mid-sequence, attention calibration consistently offers improvements from 6-15 pp. Notably, we see that attention calibration's performance curve lies almost entirely above the vanilla baseline curve (except 2 out of 24 cases), validating the effectiveness of our method in improving models' long context utilization.

\subsection{Attention calibration in practice}
In practice, to avoid the lost-in-the-middle effect, one commonly adopted workaround is to reorder the document positions, where documents considered more relevant are placed towards the beginning (or end) of the input. While these methods have led to performance improvements over the baseline without reordering, without handling the model's intrinsic bias, reordering-based methods' performance relies heavily on the correct ranking of the documents.
We are thus interested in validating whether attention calibration can be applied on top of re-ordering methods to provide another layer of improvements.

\paragraph{Attention calibration improves existing RAG pipelines.}
We continue using NaturalQuestion and SynthWiki for evaluation. We compare to existing reordering methods including:
\begin{itemize}
    \item Prompt reordering~\citep{sun2023chatgpt,liang2023holistic}: Reorder documents based on relevance score generated through prompting.
    \item \llmlinguafull~\citep{jiang2023longllmlingua}: Reorder documents using query generation as the reranking metric.
    \item Attention sorting~\citep{peysakhovich2023attention}: Reorder documents using vanilla model attention assigned to the documents.
\end{itemize}

\begin{figure*}[t!]
    \centering
    \includegraphics[width=\linewidth]{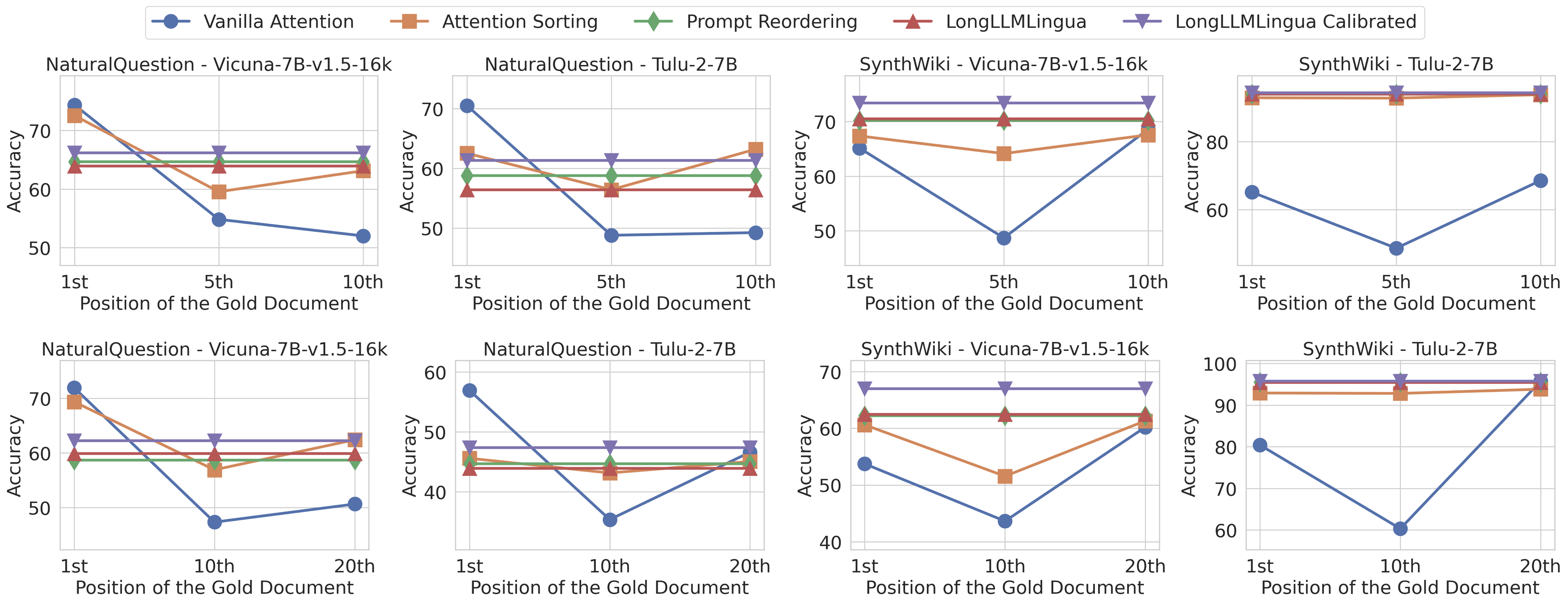}
    \caption{\textbf{Attention calibration can be applied on top of reordering-based methods to provide further performance boost. This suggests that mitigating attention bias can more fundamentally improve models' context utilization, offering a complementary way to further improve existing RAG pipeline.} Top/Bottom row: 10/20-doc. Numbers shown in Table~\ref{table:rag_table}.}
\label{fig:reordering}
\end{figure*}

In Figure~\ref{fig:reordering}, we note that \llmlingua\ and prompt reodering are invariant to the gold document's position since they compute the relevance of each document independently. First, we see that reordering methods do alleviate lost-in-the-middle problem where models' performances increase when gold documents is placed mid-sequence. More importantly, we see that by applying attention calibration on top of a reordering mechanism (\llmlingua in this case), \llmlingua\ with calibration consistently achieve the highest performance across datasets and models. These results suggest that attention calibration can more fundamentally improve models' context utilization, providing a complementary way to re-ordering methods to further improve current RAG pipeline.

%% file: sections/05-related_work.tex
\section{Related work}

\paragraph{Retrieval augmented generation.}
While LLMs exhibit strong capabilities~\citep{team2023gemini, chatgpt, touvron2023llama}, their knowledge is inherently limited in its pretraining data, and they are observed to struggle in handling knowledge intensive tasks~\citep{petroni2020kilt}. To tackle this, retrieval augmented generation (RAG) is an effective framework that retrieves relevant information from external knowledge sources to aid and ground language models' generation~\citep{lewis2020retrieval, Khandelwal2020Generalization, borgeaud2021improving, izacard2021leveraging, izacard2022few}.

Although RAG has powered many recent language model applications from question-answering~\citep{izacard2021leveraging} to automatic task completion~\citep{shen2023hugginggpt}, recent work show that LLMs tend to \textit{lost-in-the-middle}, significantly hindering the full potential of RAG~\cite{liu2023lost}.
In this work, we take a step further to understand the lost-in-the-middle problem from the viewpoint of attention bias. Moreover, we propose a remedy through attention calibration, which improves upon existing RAG frameworks.

\paragraph{Long-context utilization in language models.}
There is a rich literature on enabling LLMs to handle longer input contexts, including designing efficient training and finetuning schemes~\citep{dao2022flashattention, li2023lightseq, longchat2023, shi2023context} and inference-time methods that extend an LLM's context length~\citep{press2021train, ratner-etal-2023-parallel, xiao2023streamingllm, bertsch2023unlimiformer}.
Nonetheless, even models specifically trained for long-context suffer lost-in-the-middle problem~\citep{liu2023lost, longchat2023}.

To improve LLMs' performance on handling long contexts, recent methods design better prompting techniques and pipelines that mechanically work around the lost-in-the-middle problem~\citep{chen2023walking, jiang2023longllmlingua, peysakhovich2023attention, junqing2023never}. For instance, to avoid having the models process long input contexts, \citep{chen2023walking, junqing2023never} proposes to split long inputs into shorter contexts for models to better understand. To avoid relevant context being missed by the model, ~\citep{jiang2023longllmlingua, peysakhovich2023attention} proposes to rank the relevance of different parts of the input and re-order the most important parts to either the beginning or end of the entire input, where the models tend to focus more.

While these existing solutions lead to improved model performances by manipulating the input contexts, they do not fundamentally improve LLMs' underlying long-context utilization capability.
In contrast, we set out to directly improve LLMs' long-context utilization capability to mitigate lost-in-the-middle problem.

\paragraph{Self-attention and attention bias.}
The attention mechanism is initially introduced in RNN-based encoder-decoder architectures~\citep{bahdanau2015neural, luong2015effective}.
Building upon the self-attention mechanism, transformers~\citep{vaswani2017attention} have achieved state-of-the-art performance in various domains~\citep{devlin2018bert,dosovitskiy2020image}. Self-attention has also been widely used as a proxy to understand and explain model behaviors~\citep{clark2019does,hao2021self,vashishth2019attention}.

However, the relationship between the lost-in-the-middle problem and LLM's self-attention has been under-explored. As an initial trial, ``attention sorting''~\citep{peysakhovich2023attention} sorts documents multiple times by the attention they receive to counter lost-in-the-middle. Recently, ~\citet{he2023never} construct a dataset for training LLMs to focus on the most relevant documents among long contexts. 
Unlike the method, which necessitate significant investment in data collection and LLM tuning, our method offers an efficient solution by mitigating lost-in-the-middle problem with off-the-shelf LLMs.

%% file: sections/06-conclusion.tex
\section{Discussion}
In this work, we understand and address the lost-in-the-middle phenomenon, by establishing a connection between the phenomenon and models' positional attention bias.
We mitigate the bias by attention calibration which directly modifies the model's attention mechanism, enabling LLMs to more faithfully attend to contexts based on their relevance, rather than their position.
Experiments show that attention calibration improves the performance compared to its uncalibrated counterpart especially when relevant context occurs in the middle of the input. We additionally show attention calibration can be applied on top of existing reordering pipelines to further improve models' performance.

\section*{Limitations}

While our study presents significant advances in addressing the "lost-in-the-middle" problem and improving RAG performance in LLMs, several limitations are noteworthy:

\paragraph{Simplification of the mechanism behind positional attention bias.}
We proposed a simple hypothesis to model the positional attention bias, as shown in Eq.~\ref{eq:main-hypothesis}. However, the intrinsic mechanisms that drive this bias could be more intricate and dynamic than our current model accounts for. It is possible that some aspects of attention bias are learnable or adaptive, responding to subtle aspects of the data or training process that our current approach does not consider.

\paragraph{Computational overhead.}
Our method of calibrating positional attention bias, while effective, introduces additional computational overhead. Specifically, we require extra $O(K)$ model forward passes to calibrate attention at each position, compared to vanilla model generation. However, in this study we aim to discover and calibrate the positional attention bias from a scientific perspective. We expect that our discovery can enable future research into developing more calibration methods with lower computational overhead.  

\paragraph{Positional attention bias may be beneficial.}
Our method aims to completely remove positional attention bias. However, it is important to note that this positional bias might actually be beneficial in certain contexts. In some specific tasks or scenarios, the natural tendency of models to focus more on the beginning and end of inputs could align well with the structure of the task or the nature of the data. Therefore, understanding the tasks and the applications is required before adopting our proposed calibration method.

\paragraph{The root cause of attention bias is unclear.}
In this work, we aim to discover and understand the connection between the lost-in-the-middle problem and LLMs' intrinsic attention bias.
However, our work does not definitively pinpoint the root cause of attention bias in LLMs. The cause of such a bias could be attributed to the distribution of pretraining corpora, the transformer model architecture, and the optimization process. Future research needs to delve deeper into the origins of this phenomenon.

\section*{Ethical Statement}

In our research, we focus on enhancing the performance of large language models using existing public datasets, ensuring that no personal or sensitive data was collected or utilized. Our attention calibration method is aimed at improving the efficiency and accuracy of retrieval-augmented generation, with potential benefits across various domains including search engines, question-answering systems, and other text-based applications.
It is important to acknowledge that as our technique builds upon pre-trained language models, it may inadvertently inherit and propagate existing biases inherent in these models. Apart from this significant concern, we do not identify any other immediate risks arising from the methodologies or findings presented in our paper.

%% file: sections/07-appendix.tex
\section{Multi-doc QA datasets}
\label{sec:datasets}
We use NaturalQuestions~\citep{kwiatkowski2019natural}\footnote{\url{https://github.com/google-research-datasets/natural-questions}} (released in Apache-2.0 license) and SynthWiki~\citep{peysakhovich2023attention}\footnote{\url{https://github.com/adamlerer/synthwiki}} to conduct the experiments. Both datasets contains question-answer pairs, a gold document contains the answer, and $K - 1$ distractor documents, where $K = 10$ and $20$. 

The NaturalQuestions dataset is the subset with 2655 queries selected by \citet{liu2023lost}\footnote{\url{https://github.com/nelson-liu/lost-in-the-middle}} where the annotated long answer is a paragraph. The $k - 1$ distractor passages are Wikipedia chunks retrieved by Contriever~\citep{izacard2022unsupervised} that are most relevant to the query but do not contain any of the annotated answers in  NaturalQuestions. The distractor documents are presented in the context in order of decreasing relevance.

The SynthWiki dataset~\citep{peysakhovich2023attention} is a synthetic multi-doc QA dataset with 990 entries. All the documents in SynthWiki are GPT-4 generated Wikipedia paragraphs for fictional people, thus it can minimize the knowledge contamination issue from pre-training and ensure the LLMs can only use information from the provided context. The distractor documents are randomly sampled and randomly ordered in SynthWiki.

NaturalQuestions is collected from public English Wikipedia articles and SynthWiki is collected by GPT-4 automatic generation of English fake Wikipedia articles. These two dataset should not contain any information that names or uniquely identifies individual people or offensive content.
We ensure that the use of these two datasets was consistent with their intended purpose for academic research and in accordance with their specified licensing agreements.

\section{Implementation details}
\label{sec:exp_detail}
In our experiments, we utilize \texttt{tulu-2-7b} and \vicunafull\ as the base models. Both models consist of 32 decoder layers, each with 32 attention heads. In applying attention calibration method to intervene model attention, we apply only to the last 16 decoder layers (and all of their attention heads). We find that intervening early layers may lead to unstable generation. We leave finding the best set of attention heads to intervene as future directions~\citep{zhang2023tell}.

In the experiments, we find attention calibration to be robust to the temperature term $t$ in \eq~\ref{eq:intervene}. We set $t=5\mathrm{e}{-5}$ for all experiments.

\section{Additional experiment results}
\label{sec:addition_exp}

\paragraph{Different model formulations.}
To approximate~\eqref{eq:main-hypothesis}, in addition to linear models as shown in~\eqref{eq:add}, we also investigate log-linear models, which is defined as 
\begin{equation}
\label{eq:log-linear}
    \log \Attn(\xdoc, k) = \rel(\xdoc) + \bias(k) + \epsilon,
\end{equation}
where $\epsilon$ is a noise.
We compute rank correlation as described in Sec.~\ref{sec:calibrated_attention}. The result is shown in Table~\ref{table:add_vs_multi}. The log-linear model and linear are competitive to each other, which all result in rank correlation above $0.75$.

\begin{table}[h]
\caption{Rank correlations of linear and log-linear models.}
\label{table:add_vs_multi}
\small
\centering
\begin{tabular}{lllc}
\toprule
Model form of $f$ &  Rank correlation \\
\midrule
Linear & 0.76 \\
Log-linear & 0.75 \\
\bottomrule

\end{tabular}
\end{table}

\paragraph{Experiment tables.}
Table~\ref{table:rag_table} shows the exact numbers in our experiments. 

\begin{table*}[t]
\caption{Our proposed attention intervention by calibrated attention stably improves models' RAG performances compared to existing re-ordering based baselines.}
\label{table:rag_table}
\small
\centering
\begin{adjustbox}{width=1\linewidth}
\begin{tabular}{lll ccccc cccc}
\toprule
& & & \multicolumn{4}{c}{Gold position in 10 documents} & \multicolumn{4}{c}{Gold position in 20 documents}\\
\cmidrule(lr){4-7} \cmidrule(lr){8-11}
Dataset & Model & Method & 1st & 5th & 10th & Avg. & 1st & 10th & 20th & Avg. \\
\midrule

\multirow{12}{*}{NaturalQuestion} & \multirow{4}{*}{Vicuna} & Vanilla attention & 74.35 & 54.83 & 52.01 & 60.39 & 71.93 & 47.34 & 50.65 & 56.64\\
& & Calibrated attention & 70.84 & 62.61 & 55.78 & 63.07 & 66.40 & 56.19 & 51.75 & 58.11 \\
& & Attention sorting & 72.54 & 59.54 & 63.12 & 65.06 & 69.37 & 56.91 & 62.41 & 62.89\\
& & Prompt reordering & - & - & - & 64.63 & - & - & - & 58.68\\
& & \llmlingua & - & - & - & 63.95 & - & - & - & 59.92 \\
& & \llmlingua\ + Cal. & - & - & - & 66.17 & - & - & - & 62.22\\

\cmidrule{2-11}

 &  \multirow{4}{*}{Tulu} & Vanilla attention & 70.50 & 48.81 & 49.26 & 56.19 & 56.94 & 35.32 & 46.59 & 46.28\\
& & Calibrated attention & 71.52 & 57.13 & 63.54 & 64.06 & 57.17 & 43.08 & 61.5 & 53.91\\
    & & Attention sorting & 62.52 & 56.43 & 63.2 & 60.71 & 45.57 & 43.12 & 45.04 & 44.57\\
& & Prompt reordering & - & - & - & 58.77 & - & - & - & 44.64 \\
& & \llmlingua & - & - & - & 56.39 & - & - & - & 43.90   \\
& & \llmlingua\ + Cal. & - & - & - &  61.31 & - & - & - & 47.34 \\

\midrule

\multirow{12}{*}{SynthWiki} & \multirow{4}{*}{Vicuna} & Vanilla attention & 65.15 & 48.68 & 68.58 & 60.80 & 53.73 & 43.63 & 60.20 &	52.52\\
& & Calibrated attention & 68.58 & 53.83 & 74.14 & 65.52 & 57.77 & 51.21 & 68.78 & 59.25\\
& & Attention sorting & 67.37 & 64.14 & 67.57 & 66.36 & 60.60 & 51.55 & 61.31 & 57.82\\
& & Prompt reordering & - & - & - & 70.20 & - & - & - & 62.22 \\
& & \llmlingua & - & - & - & 70.50 & - & - & - & 62.42 \\
& & \llmlingua\ + Cal. & - & - & - & 73.43 & - & - & - & 66.96 \\

\cmidrule{2-11}

 &  \multirow{4}{*}{Tulu} & Vanilla attention & 92.22 & 81.51 & 94.34 & 89.35 &  80.40 & 60.30 & 95.75 & 78.81\\
& & Calibrated attention & 92.92 & 87.77 & 95.25 & 91.98 & 82.22 & 75.15 & 96.14 & 84.50 \\
& & Attention sorting & 92.92 & 92.82 & 93.83 & 93.19 & 94.04 & 93.53 & 95.05 & 94.20 \\
& & Prompt reordering & - & - & - & 94.04 & - & - & - & 95.55 \\
& & \llmlingua & - & - & - & 94.04 & - & - & - & 95.45 \\
& & \llmlingua\ + Cal. & - & - & - &  94.44 & - & - & - & 95.75 \\

\bottomrule
\end{tabular}
\end{adjustbox}
\end{table*}

\section{Compute and inference details}

In the experiments, we use the Huggingface Transformer package\footnote{\url{https://github.com/huggingface/transformers}} with the two models: 
Tulu-2-7B\footnote{\url{https://huggingface.co/allenai/tulu-2-7b}} and 
Vicuna-7B-v1.5-16k\footnote{\url{https://huggingface.co/lmsys/vicuna-7b-v1.5-16k}} both contains 7B parameters.
We run the experiments with two NVIDIA A100 GPUs. The inference time is roughly 1 to 3 hours on both datasets. We run our experiments with all greedy decoding without any non-deterministic factor, so we only need to run the experiments for once. Our method is a pure inference method, so there is no need to do training or hyperparameter searching.